# TEXT-BASED DETECTION OF ON-HOLD SCRIPTS IN CONTACT CENTER CALLS

Dmitrii Galimzianov and Viacheslav Vyshegorodtsev

DeepSound.AI, Kazakhstan

## ABSTRACT

*Average hold time is a concern for call centers because it affects customer satisfaction. Contact centers should instruct their agents to use special on-hold scripts to maintain positive interactions with clients. This study presents a natural language processing model that detects on-hold phrases in customer service calls transcribed by automatic speech recognition technology. The task of finding hold scripts in dialogue was formulated as a multiclass text classification problem with three mutually exclusive classes: scripts for putting a client on hold, scripts for returning to a client, and phrases irrelevant to on-hold scripts. We collected an in-house dataset of calls and labeled each dialogue turn in each call. We fine-tuned RuBERT on the dataset by exploring various hyperparameter sets and achieved high model performance. The developed model can help agent monitoring by providing a way to check whether an agent follows predefined on-hold scripts.*

## 1. INTRODUCTION

On-hold time is one of the important metrics for call centers to track [1]. Analytical companies and software providers have written many blog posts about average hold time. Some claim that on-hold time affects customer satisfaction [2,3], and others estimate the maximum allowable on-hold time and offer strategies to reduce it [4–6]. Others advise scripts containing polite phrases that agents can say when putting clients on hold [7–9].

Many papers have explored clients' behavior and satisfaction when waiting in a queue or put on hold. In [10], the authors investigate customers' satisfaction with telephone wait times. Furthermore, papers [11] and [12] analyze the decision-making process of callers in call centers. According to [13], the abandonment behavior of callers depends on their contact history. The authors of [14] develop an estimator to analyze customer patience behavior, while paper [15] analyzes consumer waiting behaviors in online communities. The authors of [16] propose using personalized priority policies for clients based on past waiting times, and the authors of [17] explore how delayed information may impact customers' behavior, demonstrating the influence of proper announcements. The amount of research dedicated to this topic shows that it is an important issue. To mitigate bad customer experience, companies need to provide specific instructions for agents on how to put clients on hold, announce the waiting time, and thank them after returning.

Natural language processing (NLP) techniques have been increasingly applied in contact center tasks [18]. In this study, we developed an NLP model to detect hold scripts in calls provided by customer service and transcribed using ASR [19]. This model can aid agent monitoring when synchronized with internal system information about the timestamps of a hold. Specifically, the model can determine the following:

- Whether the recommended scripts were said before and after a registered hold.
- Whether a hold associated with detected scripts was registered in the system.

## 2. RELATED WORK

The task of detecting hold scripts can be formulated as a key phrase detection problem. In this section, we analyze the instruments used by researchers to solve this problem in domains close to contact center calls.

In [20], the authors apply many NLP techniques to detect various keywords and key sentences in call transcriptions, including comparing words in a call with predefined words for slang, banned words, and name detection; determining cosine similarity between sentences in a call and predefined sentences for greeting, closing, and warring sentences; and examining n-gram similarity for counting the repetition of sentences in a call. Paper [21] presents a convolutional neural network trained to classify call transcripts into four classes. The authors of [22] develop a system that classifies incoming calls into predefined classes based on keywords. In the first step, ASR is applied to a call. In the second step, keywords are extracted from the text and converted into Word2Vec embeddings. Finally, keywords are passed to the RNN-LSTM trained on more than one thousand examples.

Paper [23] uses the NLU model to classify each utterance of a transcribed call into a finite set of labels. In [24], the authors construct a system for children with special needs that detects insulting and harmful speech in the context of a dialogue. They classified such speech into three types of sentences: insulting, harmful, and neutral. The authors investigate the impacts of various text-based and audio-based representations and ML classifiers in depth. The fine-tuned BERT showed the best results among all text representations. Paper [25] proposes various approaches for using BERT for text classification tasks.

Alternatively, key phrases can be directly detected in audio. In this case, the task would be formulated as a keyword spotting problem [26]. Paper [27] extensively reviews related methods. Moreover, both textual and audio features can be used, as described in [24,28–30].

## 3. MATERIALS AND METHODS

This section is organized as follows. In the first three subsections, we describe the dataset, machine learning model, and metrics used in this study. In the fourth subsection, we show how all three components contribute to the training and evaluation processes.

### 3.1. Data

We first explain the nature of the dataset, then reveal the labeling process, and finally present some quantitative characteristics.

### 3.1.1. Origin and Structure

We evaluated calls between a customer and an agent provided by a Russian telecom company, with each call containing two channels and stored in WAV format. Our in-house ASR, trained on Russian speech, was used to transcribe calls and obtain timestamps. ASR produces a CSV table, with each row corresponding to an interval with continuous speech detected by VAD. Each row can be interpreted as turn-taking in a dialogue and contains a recognized phrase and information about the start and end of the speech interval. This construction does not guarantee that a hold script is a whole phrase taking up one row in a table. Instead, a script can be part of one phrase, or it can be split into several consequent phrases. Hence, the task is not to find a script but to find a phrase containing a script.

### 3.1.2. Labeling

We formulated the task of finding phrases containing hold scripts in a call as a multiclass classification problem with three mutually exclusive classes. Annotators read each transcription and labeled each row turn as follows:

- 0 / irrelevant phrase – if a phrase does not contain hold scripts
- 1 / opening phrase – if a phrase contains a script putting a client on hold
- 2 / closing phrase – if a phrase contains a script returning to a client

Both script types are defined as widely as possible to demonstrate a general solution and obtain more positive labels. We do not require scripts to follow any additional rules (e.g., the use of polite words or announcements about the waiting time).

### 3.1.3. Statistics

The dataset consists of 1245 calls containing 37,297 text rows (phrases). The number of text rows per call varies, as shown in Figure 1. The dataset has a strong class imbalance, as both classes of interest—opening and closing phrases—are minority classes. This pattern was expected since an average call does not contain many hold scripts. We performed 10-fold stratification using the "sklearn" tool [31]. Table 1 presents the number of text rows per class in total and across folds. In addition, we wanted to investigate how the number of hold scripts changed across calls. Table 2 shows how many calls contain X (rows) and Y (columns) opening and closing scripts, respectively. The number of words per text row across classes varies, as shown in Figure 2.

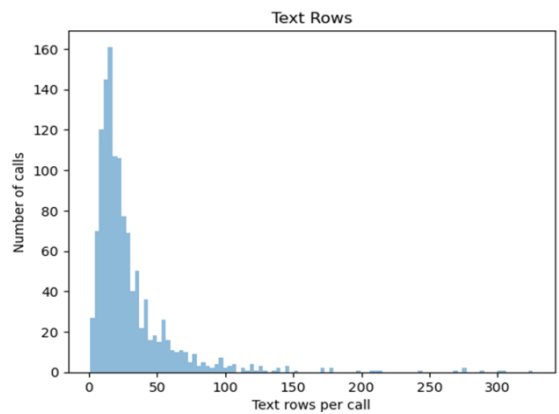

Figure 1. Number of text rows per call.

Table 1. Number of text rows per class.

| Label | 0 | 1 | 2 | 3 | 4 | 5 | 6 | 7 | 8 | 9 | Total |
|---|---|---|---|---|---|---|---|---|---|---|---|
| Irrelevant phrase | 3618 | 3648 | 3660 | 3664 | 3660 | 3661 | 3665 | 3666 | 3667 | 3667 | 36576 |
| Opening phrase | 45 | 45 | 46 | 46 | 46 | 47 | 47 | 47 | 47 | 47 | 463 |
| Closing phrase | 25 | 25 | 25 | 26 | 27 | 27 | 26 | 26 | 26 | 25 | 258 |

Table 2. Calls containing various quantities of hold scripts. Rows reflect the number of opening phrases in a call. Columns reflect the number of closing phrases in a call.

| | 0 | 1 | 2 | 3 | 4 | 5 | 6 |
|---|---|---|---|---|---|---|---|
| 0 | 891 | 25 | 3 | 1 | 0 | 0 | 0 |
| 1 | 129 | 118 | 11 | 2 | 0 | 0 | 0 |
| 2 | 13 | 5 | 5 | 1 | 0 | 0 | 0 |
| 3 | 1 | 4 | 1 | 1 | 2 | 0 | 0 |
| 4 | 5 | 1 | 7 | 1 | 1 | 0 | 0 |
| 5 | 1 | 1 | 3 | 1 | 0 | 0 | 0 |
| 6 | 0 | 1 | 2 | 0 | 1 | 0 | 0 |

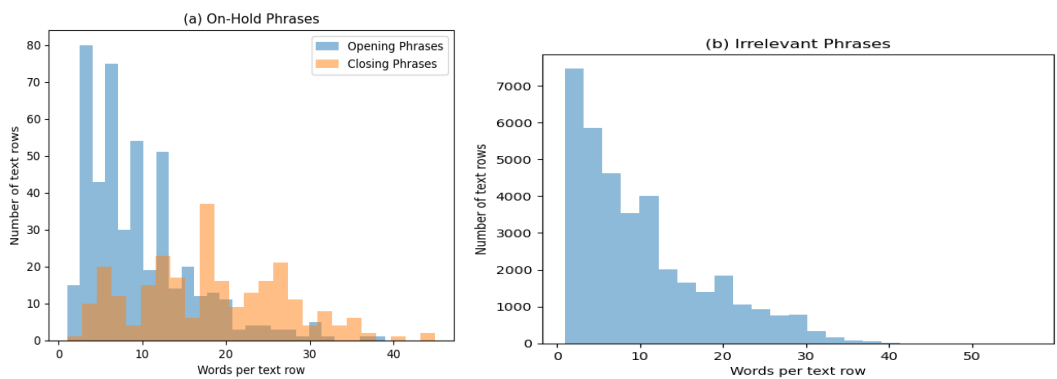

Figure 2. Number of words per text row across classes for (a) opening and closing phrases and (b) irrelevant phrases.

## 3.2. Model

Following [24], we used BERT trained on the target language and fine-tuned it on our dataset using the "transformers" library [32]. We used DeepPavlov's "Conversational RuBERT" [33] because it best suits our domain. We performed straightforward fine-tuning without freezing any layers. The results can be improved by conducting additional experiments with smarter fine-tuning strategies [25].

Threshold-moving is often used in binary classification when straightforward argmax is not sufficient for converting predicted probabilities into classes [34–36]. We needed a more complex system of threshold-moving to perform multiclass classification. Considering the class imbalance, we used the following heuristics: we selected the argmax of the first and second classes only if the sum of their probabilities was greater than or equal to the predefined threshold. In other words, we selected an argmax between the opening and closing phrases if the sum of their probabilities was significant enough.

## 3.3. Metrics

We used multiclass classification metrics to evaluate the model's performance. For each experiment, we tracked multiple metrics using the "evaluate" library [37]: recall, precision, F1, accuracy, and ROC AUC. We used macro-averaging for each metric type. In one experiment, we chose the best hyperparameter set in two steps. In the first step, we computed the best F1-macro for a given hyperparameter set as follows:

- First, we selected the best checkpoint across the training process based on ROC AUC (macro average and OVR multiclass) since it does not require a threshold.
- Second, for the selected checkpoint, the optimal threshold value is automatically adjusted based on F1-macro.

In the second step, we selected the best hyperparameter set based on its F1-macro.

## 3.4. Training and Evaluation

We trained models for five epochs using the default AdamW optimizer with a weight decay equal to 0.01 and the default scheduler [38]. We set the batch size at 16, and, given the distribution of phrases' lengths (see Figure 2), we set the maximum length in the tokenizer at 128. As a starting point, we set the learning rate at 5e-6.

We set aside one of the 10 folds as a test fold. We then performed nine-fold cross-validation, which involved training on eight folds and adjusting the threshold on the ninth fold. Given one hyperparameter set, we had nine best (based on ROC AUC) checkpoints due to cross-validation. We then set one shared threshold value for the checkpoints using the following algorithm:

- Predict the probabilities for each fold using associated checkpoints.
- Concatenate predictions for all nine folds.
- Get all unique threshold values as the sum of the probabilities of the first and second classes for each prediction.
- For each unique threshold value:
  - Compute the metric for each fold.
  - Compute the final metric as the average of the nine metric values.
- Choose the best threshold value based on the final metric.

We made predictions for the test dataset using each of the nine checkpoints, but one shared best threshold value, resulting in nine metric values. We reported the final metric of the test dataset as the average of nine values.

# 4. RESULTS

We conducted two experiments to adjust two important hyperparameters: class weights and learning rate. First, we determined the best value for class weights, assuming that class weight is a significant parameter, given the strong class imbalance in the dataset. We attempted to diminish the influence of the dominant class of irrelevant phrases by reducing its weight. Table 3 contains the metrics for various class weights. Second, we found the best value for the learning rate and set the class weight value to the best option as determined in the previous experiment. Table 4 contains the metrics for various learning rates. We found that the best class weight value is [0.075, 1, 1] and the best learning rate value is 3e-6. Figure 3 depicts the loss and metrics logged while training the best model. The final F1-macro on the test dataset equals 0.9014.

Table 3. Experiment 1. Class weight adjustment.

| Class weight | ROC AUC | Best threshold | Recall | Precision | Balanced Accuracy | F1 |
|---|---|---|---|---|---|---|
| [0.005, 1.0, 1.0] | **0.9946** | 0.5449 | **0.9135** | 0.8529 | **0.9135** | 0.8778 |
| [0.01, 1.0, 1.0] | 0.9942 | 0.9791 | 0.8824 | 0.8899 | 0.8824 | 0.8836 |
| [0.02, 1.0, 1.0] | 0.9935 | 0.5092 | 0.9095 | 0.8666 | 0.9095 | 0.8834 |
| [0.05, 1.0, 1.0] | 0.9939 | 0.9662 | 0.8802 | 0.9002 | 0.8802 | 0.8876 |
| [0.075, 1.0, 1.0] | 0.9931 | 0.9627 | 0.8805 | **0.9063** | 0.8805 | **0.8908** |
| [0.1, 1.0, 1.0] | 0.9931 | 0.9128 | 0.8883 | 0.8927 | 0.8883 | 0.889 |
| [1.0, 1.0, 1.0] | 0.9943 | 0.3347 | 0.9052 | 0.8706 | 0.9052 | 0.8843 |

Table 4. Experiment 2. Learning rate adjustment.

| Learning rate | ROC AUC | Best threshold | Recall | Precision | Balanced Accuracy | F1 |
|---|---|---|---|---|---|---|
| 5.e-7 | 0.9858 | 0.5541 | 0.8126 | 0.8504 | 0.8126 | 0.8185 |
| 1.e-6 | 0.9925 | 0.8656 | 0.8877 | 0.8985 | 0.8877 | 0.8921 |
| 3.e-6 | **0.9947** | 0.8422 | **0.9113** | 0.8949 | **0.9113** | **0.9014** |
| 5.e-6 | 0.9931 | 0.9627 | 0.8805 | **0.9063** | 0.8805 | 0.8908 |

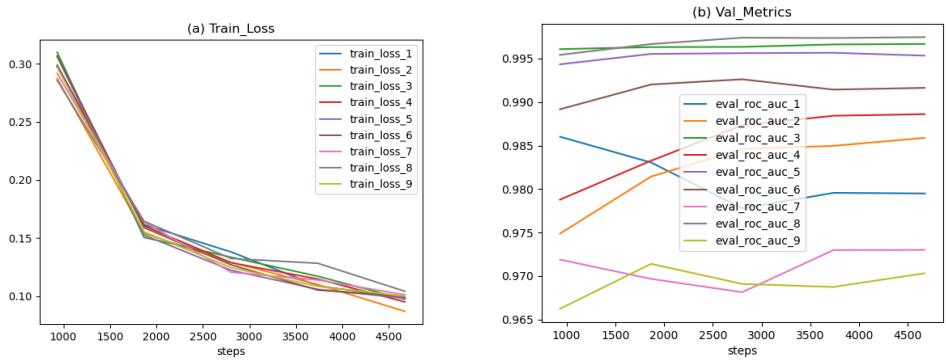

Figure 3. The training process of the best model: (a) loss on the train set and (b) metrics on the validation set.

## 5. DISCUSSION

The detection method described in the paper can be extended to all other types of keywords or sentences spoken in a call center conversation that need to be tracked: namely, greetings and closing sentences, slang words, and banned words.

The algorithm below describes how the detection model can be applied. We assume that there is a system that stores audio calls and on-hold timestamps for each call. For any given call, the following steps can be taken:

- Produce a transcription by ASR.
- For each phrase-turn in the transcription, use the developed model to predict whether a phrase is the start of a hold, the end of a hold, or irrelevant to a hold.
- Retrieve any on-hold timestamps from the system and match them to the timestamps of phrases.
- Check whether any phrase belongs to the relevant predicted class among the phrases spoken just before the start of a hold. If there is no such phrase, the agent did not follow the script when putting a client on hold. The same checking procedure can be applied to the end of a hold.
- Check whether any phrases are classified as opening or closing but are not matched to any hold timestamps registered in the system. Such phrases indicate that an agent asked a client to wait but did not register a hold.

Agents' failure to follow scripts can negatively affect customer experience. If many unregistered holds occur, the average hold time metric may not adequately reflect the situation.

## 6. CONCLUSIONS

We developed an NLP model to detect hold scripts in customer service telephone dialogues transcribed using ASR. We collected transcriptions of in-house calls and constructed a dataset using a dialogue turn as a unit. We manually labeled the dataset's phrases into three classes: putting a client on hold, returning to a client, and phrases irrelevant to hold scripts. We fine-tuned RuBERT on the dataset and used the threshold-moving technique. The model yielded high results despite a strong class imbalance. Additionally, we improved the model's performance through hyper parameter optimization by adjusting class weight and learning rate. The best model showed an F1-macro equal to 0.9014 on the test dataset, indicating the model's capability to be used in production. When used in combination with hold timestamps, the developed model can enhance agent monitoring by checking whether agents follow predefined hold scripts. The described method can be extended to detect any other important key phrases, such as greeting scripts or banned words.

## AUTHOR CONTRIBUTIONS

Dmitrii Galimzianov: conceptualization, methodology, investigation, software, formal analysis, visualization, writing—original draft preparation. Viacheslav Vyshegorodtsev: resources, data curation, writing—review editing. All authors reviewed the results and approved the final version of the manuscript.

## AVAILABILITY OF DATA AND MATERIALS

The in-house dataset of transcribed calls used in this study cannot be shared due to privacy issues. The training and evaluation code is available in our public GitHub repository: https://github.com/gal-dmitry/HOLD_DETECTION_PUBLIC.

## CONFLICTS OF INTEREST

The authors declare that they have no conflicts of interest to report regarding the present study.